\pdfoutput=1

\documentclass[11pt]{article}

\usepackage[]{acl}

\usepackage{times}
\usepackage{latexsym}
\usepackage{multirow}

\usepackage[T1]{fontenc}

\usepackage[utf8]{inputenc}

\usepackage{microtype}
\usepackage{graphicx}
\usepackage{enumitem}
\usepackage{hyperref}
\usepackage{booktabs} 
\title{The Uli Dataset: An Exercise in Experience Led Annotation of oGBV}

\author{Arnav Arora$^{1}$, Maha Jinadoss, Cheshta Arora, Denny George$^{2}$, Brindaalakshmi$^{7}$, \\ \textbf{Haseena Dawood Khan$^{3}$, Kirti Rawat, Div, Ritash, Seema Mathur$^{4}$, Shivani Yadav$^{8}$,}\\ \textbf{ Shehla Rashid Shora$^{5}$, Rie Raut, Sumit Pawar, Apurva Paithane, Sonia, Vivek,}\\ \textbf{ Dharini Priscilla, Khairunnisha$^{3}$, Grace Banu, Ambika Tandon$^{6}$},\\ \textbf{Rishav Thakker, Rahul Dev Korra, Aatman Vaidya$^{2}$, Tarunima Prabhakar$^{2}$  \thanks{\emph{\small~For any questions about this paper please email Tarunima Prabhakar at tarunima@tattle.co.in}}} \\
$^{1}$University of Copenhagen, Denmark,  $^2$Tattle Civic Tech, $^{3}$Bebaak Collective, \\ 
$^{4}$National Council of Women Leaders, $^{6}$ Center for Internet and Society,\\ $^{7}$Independent, $^{8}$Chambal Media/Khabar Lahariya \\
{\small $^1$\texttt{aar@di.ku.dk}} {\small $^2$\texttt{\{denny,aatman,tarunima\}@tattle.co.in}}\\{\small $^{5}$\texttt{shehla.shora@gmail.com}, $^{8}$\texttt{shivaniyadav48@yahoo.com}}
}

\begin{document}
\maketitle
\begin{abstract}
Online gender-based violence has grown concomitantly with the adoption of the internet and social media. Its effects are worse in the Global majority where many users use social media in languages other than English. The scale and volume of conversations on the internet have necessitated the need for automated detection of hate speech and, more specifically, gendered abuse. There is, however, a lack of language-specific and contextual data to build such automated tools.
In this paper, we present a dataset on gendered abuse in three languages- Hindi, Tamil and Indian English. The dataset comprises of tweets annotated along three questions pertaining to the experience of gender abuse, by experts who identify as women or a member of the LGBTQIA+ community in South Asia. Through this dataset, we demonstrate a participatory approach to creating datasets that drive AI systems. 
\end{abstract}

\section{Introduction}
Internet adoption promises connectivity, economic opportunity, and political agency. But for women and members of the LGBTQIA+ community, the internet and, in particular, social media can be a site of harassment and targeting. Some surveys put the incidence of online gender-based violence (oGBV) at over 50\% ~\cite{Hicks2021}. Nearly 85\% of women have seen violence against women online~\cite{economist}. The most common site for such encounters is social media platforms. oGBV is now seen as an extension of offline violence, with its effects being worse ``in countries with long-standing or institutionalized gender inequality'' \emph{(ibid)}. A study found that the volume of misogynistic Facebook posts and tweets, as well as individuals' engagement with them, spiked during lockdowns in the pandemic, with a 168-percent increase from the same period in 2019 ~\cite{unwomen}. The prevalence of oGBV restricts people from marginalized genders from accessing economic, social and political opportunities, threatening to exacerbate the digital divide.

As with hate speech, tackling gendered abuse online at scale necessitates automated approaches to detect it. Such approaches depend on language and context-specific datasets, which are sparse beyond English and a few other languages. With the goal of addressing oGBV in the majority of the world, and more specifically in India, we focused on creating a dataset of gendered abuse from India. We further recognized the importance of centering the lived experience of abuse in data work~\cite{DIgnazio2020}. While this project extends prior work on crowd-sourced annotations of hate speech, it is distinct in attempting to source these annotations from expert annotators, i.e. activists and researchers who have encountered or responded to online abuse. As described in the next section, this annotation was carried out as a part of a project on user-end interventions to protect oneself and respond to oGBV. Machine learning driven redaction of tweets, for which this dataset was created, was just one feature. Thus, we started this exercise from the primary position of- what constitutes gender abuse? This makes our work distinct from several other datasets that use gender as one of many axis on which hate speech is expressed~\cite{Trac_KUMAR}. While gendered abuse inevitably overlaps with hate speech, by starting with the question of what is specifically gendered abuse in social media discourse, we are able to describe the experience in more detail. 

We started this data collection exercise by recognising that any attempt to capture oGBV in a dataset will necessarily involve simplifications and omissions. A dataset cannot capture the whole experience of oGBV that involves a number of behaviours such as trolling, non-consensual sharing of private information and repeated unwanted engagement. Furthermore, oGBV is often an extension of offline violence. Online experiences are overlaid on offline socio-economic vulnerabilities and intersected identities to produce a specific experience of violence. This is a background context that cannot be captured in a dataset. This dataset captures a very small aspect of the experience of oGBV- that which is patently visible in text-based statements. Within this narrow scope, we use the terms oGBV and gendered abuse interchangeably. In the annotation guideline, we used the term gendered abuse instead of oGBV. 

This dataset, inspired from values of feminist technologies such as inclusion, intersectionality and care, is an attempt at participatory models of machine learning development \cite{Clancy2021Introduction}. The definition of gendered abuse, as well as the annotations, came from activists and researchers who identify as a marginalized gender and have encountered or responded to online or offline abuse. This paper describes the process of creation of the dataset in three Indian languages: Hindi, Tamil and Indian English.




\section{Background}
As with all datasets on abuse detection, our dataset too had to contend with the social and theoretical task of defining abuse~\cite{vidgen-etal-2019-challenges}. This dataset was created as a part of a larger project to build a browser-based tool\footnote{\url{https://uli.tattle.co.in/}} to help mitigate the effects of online gender-based violence on those who are at the receiving end of it. The tool includes a machine learning driven feature for the redaction of content as well as non-machine learning features such as the redaction of problematic words and tools for archiving. The tool aimed to center the experiences of those at the receiving end of oGBV. Through formal and semi-structured interviews and focus group discussions with over thirty activists and researchers working on gender and minority rights in South Asia, we identified the varieties of ways in which harm was manifested and perceived in this group. The interviews and focus group discussions were conducted over Zoom from July 2021 to October 2021. The discussions emphasized the contextual nature of online gender and sexual abuse. Participants were concerned with who made a statement, to whom it was directed and the ongoing global and local events when the post was written. 
 
The location of moderation - 'user-end' as opposed to platform-end - shaped the respondent's views on how harm and abuse should be understood. First, participants in the qualitative research phase did not express concerns about excessive moderation through automation. Instead, participants mentioned that from the perspective of mitigating harm to the person harassed, it is acceptable if the machine learning model `over' moderates on certain classes of speech, such as hate speech. Second, they mentioned that the model should be able to capture instances that escape platform-centered moderation because they don't violate community guidelines. Thus, oGBV or gendered abuse, as defined in this dataset, may be broader than other datasets. 

The focus group discussions surfaced that a large proportion of abuse was in the form of images and videos. In this first attempt to build a survivor-centered dataset in Indian languages, the scope was oGBV as manifested in text-based abuse. We seek to address this limitation in future iterations of our project

\section{Related Work}
Abusive speech has been studied under several overlapping categories \cite{waseem-etal-2017-understanding} such as hatespeech ~\cite{badjatiya_2017, waseem-2016-racist,    Davidson_2017}, offensive language \cite{chen_2012, Nobata} and trolling \cite{mojica-2018-modeling}. \citet{comma-v01} and \citet{waseem-hovy-2016-hateful} specifically focus on gender bias and sexism, respectively, within hate speech. As described in section \ref{sec:guideline}, we tested the categorizations proposed in these papers to understand our data better and develop our annotation guideline. ~\citet{waseem-2016-racist}'s dataset of hate speech on Twitter is especially relevant to our work since they also relied on feminist activists and showed that systems trained on these expert annotations outperform systems trained
on amateur annotations.

While most of the aforementioned papers focus on English language content, there has been a push to expand abuse and hate speech detection to languages other than English. Within Indian languages, Hindi has received considerable attention.~\citet{Mandl2019OverviewOT, Mandl2020OverviewOT} proposed a dataset for hate speech in Hindi language consisting of $5K$ and $6K$ posts sourced from Twitter. Following the previous work,~\citet{mandl2020overview}  shared another hate speech dataset of $3.6K$ posts scrapped from YouTube and Twitter.~\citet{bohra2018dataset} introduced a code-mixed Hindi dataset on Hate Speech containing $4.5K$ tweets, out of which, $1.6K$ tweets are labelled hateful, and the remaining $2.9K$ are non-hateful. Tweets are annotated as hate speech or normal speech.~\citet{Saroj2020AnIL} proposed a dataset for Hindi language on offensive speech containing $2K$ posts from Twitter and Facebook.~\citet{velankar2021hate} created a dataset for Hindi and Marathi on hate and offensive speech with $4.5K$ and $2K$ posts respectively.~\citet{romim2021hate} created a dataset on hate speech consisting of $30K$ comments in Bengali from YouTube and Facebook, $10K$ comments are annotated as hateful.~\citet{gupta2022multilingual} proposed a large-scale (150K) abusive speech dataset of comments in Hindi, Tamil, Telugu, Kannada and Malayalam sourced from ShareChat. 
~\citet{chakravarthi2021findings} created a code-mixed dataset on offensive speech consisting of YouTube comments in Kannada, Malayalam, and Tamil, with 7.7K comments for Kannada, 20K for Malayalam, and 43K for Tamil.~\citet{bhardwaj2020hostility} collect posts from Twitter and Facebook in Hindi and provide annotations for hostile posts including fake news, hate speech, and other offensive posts. The ~\citet{comma-v01} dataset specifically looked at gendered and communally charged comments in four Indian languages. The dataset was annotated at three levels: aggression, gender bias, and communal bias.

While many of these datasets are larger in size than the one we collected, none of them have survivor-focused definitions and guidelines along with expert annotations like the one we provide in our work.

\section{Corpus Creation}
To build a robust and diverse dataset, we followed a two step process. We first scraped a large collection of tweets and then selected data for annotation from that collection using a semi-supervised approach. In the first iteration of the project, we focused on three Indian languages- Hindi, Tamil and `Indian' English. Indian English~\citep{sailaja2012indian} was suggested as a distinct language by some of the activists we engaged with. It was felt that the specific way English was used in India, which included some transliteration of words from other languages and code-mixing, made it distinct enough to merit specific attention. 

\subsection{Unlabelled dataset collection}
For the initial collection of a large unlabelled dataset, we crowdsourced a list of slurs and offensive words/phrases from the group of activists and researchers. Additionally, we created a list of accounts that are often at the receiving end of hate online, as well as a list of accounts that are often found perpetuating hate and abuse on Twitter (now called X), by manually scanning conversations on the platform. This was complemented by data from ~\citet{msr_influencers} and ~\citet{IT4C}\footnote{The data was requested from IT for Change while research from this report was ongoing}, that contained a list of influential or highly active women on Twitter/X who are often at the receiving end of online abuse and harassment as well as annotated data for different variants of potential harm online. Thus, we scraped tweets using three criteria: (1) crowdsourced slurs and keywords, (2) tweets by known perpetrators, and (3) replies to highly influential women on Twitter. In total, we were able to scrape close to 1.3 million tweets from 2018-2021. We used the Python Twint library\footnote{\url{https://github.com/twintproject/twint}} to collect public tweets that matched the three criteria. We filtered for language based on the language assigned by Twitter. We replaced all user handles mentioned in the posts, as detected by a regex query of words starting with '@', with the term <handle replaced>. Thus, the experts could not see who was being addressed in a post, and whose post was the message a reply to. 

\subsection{Stratified Pooling}
Our annotation budget determined the dataset size: roughly 8000 posts in three languages. Despite our strategy of collecting data based on problematic keywords, the majority of the dataset was non-oGBV. Creating an annotation set by randomly sampling posts from the larger unlabelled dataset of 1.3 million tweets would have resulted in a very small dataset of tweets containing gendered abuse. Thus, we used stratified pooling to create a dataset in which the percentage of abuse is higher than the larger data corpus, and possibly higher than Twitter in general. To do this, we first assign noisy labels to our unlabelled dataset using democratic co-learning as done in prior work~\citep{rosenthal-etal-2021-solid}. We used various models trained on open-source datasets for related tasks of offensive language detection, misogyny detection, and hate speech detection. In Table~\ref{tab:pooling_infra}, we show the set of datasets and models used for our semi-supervised annotation per language.

We thus obtain confidence scores for the models listed above on our large unlabelled dataset. Since the models are trained on datasets pertaining to different tasks, we treat these models as Mixture-of-Experts (MoE) in their own tasks. To obtain a consensus among them, we average the confidence scores of all the models per post. We then bin the posts based on their averaged confidence scores, categorising them into 10 categories. Finally, we randomly sample a fixed number of posts from each bin to include in the final dataset used for annotation. For English and Hindi, the number of posts selected from each bin are shown in Table~\ref{tab:eng_pooling}. For Tamil, due to the lack of posts in each bin, we select posts based on two bins, as shown in Table~\ref{tab:tamil_pooling}. 
The selection of posts from these bins was made to increase diversity in the kind of content in our final dataset as well as to maintain balance among the easily identifiable hate speech by existing models and the examples on which the models disagree (which represents the posts with mean scores close to 0.5).

\begin{table}[t!]
    \centering
    \begin{tabular}{|p{1cm}|p{3.5cm}|p{1.75cm}|}
    \hline
     &  Datasets & Model Used \\
    \hline
    English &  ~\citet{hatexplain},~\citet{kumar-etal-2018-benchmarking}
                ~\citet{basile-etal-2019-semeval}, ~\citet{zampieri-etal-2019-semeval}, ~\citet{Founta} &  Twitter Roberta \\
    \hline
    Hindi &     ~\citet{bohra2018dataset}, ~\citet{bhattacharya-etal-2020-developing},~\citet{kumar-etal-2018-benchmarking}, ~\citet{hasoc} &                        Bert Based Code-mixed model\\
    \hline
    Tamil & ~\citet{chakravarthi-etal-2020-corpus},~\citet{hasoc} & Indic Bert  \\
    \hline

    \end{tabular}
    \caption{List of datasets and models used for MoE based pooling across the three languages}
    \label{tab:pooling_infra}
\end{table}

\begin{table}[]
    \begin{tabular}{ |c|c| } 
    \hline
    Toxicity score range & No. of Tweets \\
    inclusive of the extreme values & \\
    \hline
    0-0.1 & 400\\ 
    \hline
    0.11-0.2 & 800\\ 
    \hline
    0.21- 0.3 & 800\\ 
    \hline
    0.31-0.4 & 1000 \\ 
    \hline
    0.41-0.5 & 1000 \\ 
    \hline
    0.51- 0.6 & 1000 \\ 
    \hline
    0.61-0.7 & 1000 \\
    \hline
    0.71-0.8 & 800 \\
    \hline
    0.81- 0.9 & 800 \\
    \hline
    0.91-1 & 400 \\
    \hline    
    \end{tabular}
    \caption{Number of posts sampled per bin for our MoE based pooling of English and Hindi data}
    \label{tab:eng_pooling}
\end{table}

\begin{table}[]
    \begin{tabular}{ |c|c| } 
    \hline
    Toxicity score range & No. of Tweets \\
    inclusive of the extreme values & \\
    \hline
    0-0.5 & 4000 \\
    \hline
    0.51-1 & 4000 \\
    \hline    
    \end{tabular}
    \caption{Number of posts sampled per bin for our MoE based pooling of Tamil data}
    \label{tab:tamil_pooling}
\end{table}


\section{Annotation Guideline}
\label{sec:guideline}
The literature review and focus group discussions informed our early criteria for marking abuse. Four researchers in the team who identify as marginalized genders annotated posts in small batches, as per different typologies such as intersectional themes (ableist, transphobic and queerphobic, body shaming), kinds of abuse (sarcasm, threats, derogatory comments), explicit or implicit nature of abuse. This team consisted of language speakers from each of the three languages- English, Hindi and Tamil. Such granular labelling helped the team familiarize itself with the data, as well as surface disagreements within the team. Over three months, the team repeatedly annotated batches of data, distilling the initial typologies to the most essential labels and converging on a guideline to describe the purpose of the label. The simplification of labelling was essential since the labelling had to be carried out by activists and researchers with other primary commitments. We converged on the following two labels:
\begin{itemize}[noitemsep]
\item Is the post gendered abuse 
\item Does the post contain explicit or aggressive language. 
\end{itemize}

When the interrater agreement between the team members across the labels exceeded 0.3, the team opened the labelling task to the external group of gender rights activists and researchers.

We created an annotation guideline with definitions for the labels and examples. The guideline was initially written in English\footnote{\scriptsize{\url{https://docs.google.com/document/d/1JRPGCSM-9YUc0UWIyc3u7NDyvMTYxWFabsiUC6T4AcI/edit?usp=sharing}}}
and then translated into Hindi\footnote{\scriptsize{\url{https://docs.google.com/document/d/1JRPGCSM-9YUc0UWIyc3u7NDyvMTYxWFabsiUC6T4AcI/edit?usp=sharing}}} and Tamil. The examples in the Hindi and Tamil guidelines were picked by the team members speaking the language to mirror the motivation for including the corresponding examples in the English guideline. 

To onboard the annotators to the guideline, we paired annotators and asked them to annotate a hundred posts as per the guideline. Where they disagreed, we asked them to discuss their reasons for their choice of label. This exercise was repeated 2-3 times for each pair. While in some cases, the disagreement in the label was a result of misunderstanding the guideline, we also learnt that absent any context to a post, such as the relationship between the person posting and the receiver (in case of replies) or the broader conversation, each annotator assumed context. This shaped whether they perceived the post as gendered abuse or not. Thus, to reduce some of the ambiguity in the imagined context, we broke the first label into two parts:
\begin{itemize}[noitemsep]
\item Is the post gendered abuse when not directed at a person of marginalized gender?
\item Is the post gendered abuse when it is directed at a person of marginalized gender?
\end{itemize}

The first label would capture outright misogynistic comments, such as those commenting on women's capabilities to participate in professional or public life. The second label is a more expansive one that we recognize could capture all forms of abuse. From the perspective of the expert annotators, any form of hate speech, even if the terms used are not gendered, when directed at a person of marginalized genders is gendered abuse. The inclusion of the second label allowed us to accommodate for one assumption in context of the post- the gender of the person receiving the content.   

The final annotation tasks were as follows:

\begin{itemize}

\item \textit{Is this post gendered abuse when not directed at a person of marginalized gender and sexuality?} 
Posts which are not otherwise gendered, sexist, or trans-phobic but become oGBV if they are directed towards gender or sexual minorities are labelled as yes (1) for this question. This label accounts for hate speech that can be used to target gender or sexual minorities.

\item \textit{Is the post gendered abuse when directed at a person of marginalized gender and sexuality?}
This question is answered yes (1) for misogynist, sexist, trans-phobic comments, or general backlash against feminist principles, or posts that explicitly attack someone for their gender and sexual identity. 

\item \textit{Is this post explicit/aggressive?}
This question will be answered as yes (1) when posts contain slur words or aggressive language, even if intended as a jest. This question captures posts that use explicit or aggressive language, even if the totality of the post is not abusive. 

\end{itemize}

All these tasks were optional. An annotator could skip one or all questions. When skipping all questions, annotators were requested to leave a note in a free-form text field for us to understand why the post was not annotated. 



\section{Annotator profiles}
The project started with twenty annotators, but only sixteen annotators remained till the end: six for Hindi and five each for English and Tamil. For those who left the project before sufficient time for onboarding on the annotation guidelines, we discarded the annotations. 
Most annotators were individuals who were active in gender and sexual research and activism in India. One of them belongs to Sri Lanka, and some have lived in or moved to other countries during this project. They either belonged to or worked with the affected groups/communities or were themselves at the receiving end of violence and online abuse. They all self-identified and situated themselves on the LGBTQIA+ spectrum, and at least a third of them explicitly identified and situated themselves on the vulnerable religious and caste backgrounds in India—Muslim and Dalit—ensuring an intersectional approach in the annotation task. The activists and researchers represent a range of socio-cultural as well as geographical backgrounds. Each annotator was actively involved in gender and sexual rights-based activism in India and foregrounded a wide range of political perspectives in their work. The Hindi group had three senior and three early career participants. The Hindi group had a higher percentage of early-career participants. The Tamil group represented more senior and middle-aged participants and a greater transnational diversity. Across the three language groups, some annotators identified as Dalit, Trans and Muslim. A majority of the annotators came from urban centers such as Delhi, Bangalore, Pune and Chennai. All the annotators, except one who requested anonymity, are listed as co-authors on this paper.


\section{Allocation of Posts to Annotators}
We started with the goal of having a total of 8000 posts annotated in each language, with 20\% of the posts (1600) being annotated by three experts. At the beginning of the exercise seven experts signed on as annotators for English, six for Hindi and six for Tamil (nineteen total). While the experts were compensated for the task which was tied to the number of posts annotated, their engagement was considered voluntary and could be terminated whenever they wished to do so and without any contractual obligations. The posts were assigned to annotators in batches. To accommodate for drop-outs and possible drop-outs, some of the posts were allocated to more than three annotators. Consequently, all annotators were not allocated an equal number of posts. In some cases, despite the reallocation, we did not get the required number of annotations. Thus, the final dataset has fewer than 8000 posts. The total number of posts in every language where at least one label was annotated is show in Table \ref{final_tally} 

\begin{figure}[h]
    \caption{User-interface to annotate posts}
    \centering
    \includegraphics[width=0.25\textwidth]{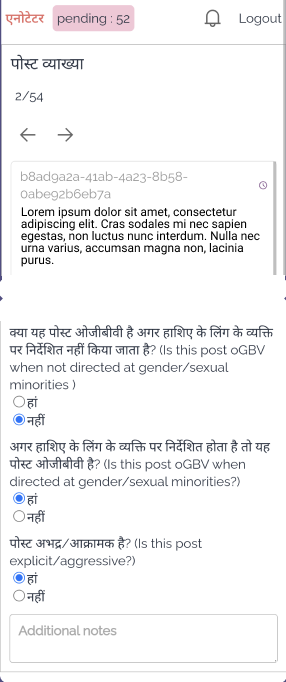}
    \label{fig:Uli UI}
\end{figure}

The annotators annotated the posts using a custom UI that we developed for this task
\footnote{\url{https://github.com/tattle-made/Uli/tree/main/annotators}}. The interface was accessible through a URL that could be opened on any browser. The UI was made responsive to enable annotations on mobile. The languages of the UI changed based on the language the annotator was working on. Figure \ref{fig:Uli UI} shows the annotation interface. The posts were annotated between March 2022 and July 2022. 

\begin{table}[]
    \begin{center}
    \begin{tabular}{ |c|c| } 
    \hline
    Language & Posts with at least \\
     &         one label annotated \\
    \hline

    English & 7638 \\ 
    Hindi & 7714 \\ 
    Tamil & 7914 \\
    \hline

    \end{tabular}
    \end{center}
    \caption{Dataset Size}
    \label{final_tally}
\end{table}



\section{Dataset Analysis}
The batch-wise allocation of tweets to annotators, some annotators dropping out and annotators skipping some labels, resulted in some posts having an even number of annotations (even if the post was allocated to an odd number of annotators). Table \ref{tab:annotation_number} shows the number of posts that were annotated by a specific number of annotators (ranging from 1 to 6). 

\begin{table*}[h]
    \begin{center}
    \begin{tabular}{|c|c|c|c|c|c|c|c|}
    \hline
              &   N   & 1    & 2   & 3    & 4   & 5  & 6 \\ \hline
    \multirow{3}{*}{English} & Label 1 & 6035 & 484 & 1112 & 7   &    &   \\
                             & Label 2 & 6035 & 484 & 1112 & 7   &    &   \\
                             & Label 3 & 6035 & 484 & 1112 & 7   &    &   \\ \hline
    \multirow{3}{*}{Hindi}   & Label 1 & 6074 & 62  & 1530 & 46  & 1  &   \\
                             & Label 2 & 6069 & 61  & 1530 & 46  & 1  &   \\
                             & Label 3 & 6075 & 61  & 1530 & 46  & 1  &   \\ \hline
    \multirow{3}{*}{Tamil}   & Label 1 & 6412 & 13  & 1086 & 349 & 48 & 6 \\
                             & Label 2 & 6411 & 13  & 1086 & 349 & 48 & 6 \\
                             & Label 3 & 6412 & 13  & 1086 & 349 & 48 & 6 \\ \hline
    \end{tabular}
    \end{center}
    \caption{Number of posts annotated by `n' number of annotators}
    \label{tab:annotation_number}
\end{table*}

We also explore the relationship between label 1: posts that are gendered abuse when not directed, and label 2: posts that are gendered abuse when directed at a person of marginalized gender. We find that in 6058 posts, at least one annotator annotated label 1 and label 2 differently. Table \ref{tab:label1_label2} shows the language-specific breakdown. The Appendix contains an annotator specific breakdown of how frequently an annotator marked label 1 and label 2 differently.

\begin{table}
\begin{center}
   \begin{tabular}{ |c|c| } 
   \hline
   Language & Number of posts \\
   \hline
    English & 1342 \\
    \hline
    Hindi & 3094 \\
    \hline
    Tamil & 1622 \\
    \hline
  \end{tabular}
\end{center}
  \caption{Posts where at least one annotator marked \\ label 1 and label 2 differently}
  \label{tab:label1_label2}
\end{table}

\subsection{Agreement Assessment}
We started with the understanding that there could be significant disagreement across the annotators on what constitutes gendered abuse. Yet, we calculate the agreement score for posts to understand the level of agreement or lack thereof. Table \ref{tab:agreement} shows the Krippendorf alpha values for the three labels for all three languages. Notably, the scores varied across the three languages, with the scores across the three labels being higher for Tamil. Tamil and Hindi have the highest agreement on when the post is explicit or aggressive.

\begin{table}[]
    \begin{center}
    \begin{tabular}{ |c|c|c| } 
    \hline
    Language & Label & Values \\
    \hline
    \multirow{3}{*}{English} & Label 1 & 0.402 \\ 
                             & Label 2 & 0.258 \\ 
                             & Label 3 & 0.35 \\ 
    \hline
    
    \hline
    \multirow{3}{*}{Hindi} & Label 1 & 0.396 \\ 
                           & Label 2 & 0.314\\ 
                           & Label 3 & 0.501 \\ 
    \hline
    
    \hline
    \multirow{3}{*}{Tamil} & Label 1 & 0.488 \\ 
                           & Label 2 & 0.411 \\ 
                           & Label 3 & 0.721 \\ 
    \hline
    
    \end{tabular}
    \end{center}
    \caption{Krippendorf Alpha}
    \label{tab:agreement}
\end{table}
%
\subsection{Known issues}
Due to an issue with allocation of posts in one of the earliest batches, a small number of posts were reassigned to the same annotators. That is, annotators were asked to label the posts that they had already labelled. While we could discard these, we retain them as they convey important information: for five posts (two in Tamil and three in Hindi), the annotators labelled the post differently in every iteration. The value of the label for these posts is a decimal that reflects the average score.

%

\begin{table}[t!]
\begin{tabular}{|c|c|c|c| } 
\hline
Language & Label & IndicBERT & XLM-T \\
\hline
\multirow{3}{*}{English} & Label 1 & 0.44 & \textbf{0.77}\\ 
& Label 2 & 0.38 & 0.70 \\ 
& Label 3 & 0.37 & 0.74\\ 
\hline
\multirow{3}{*}{Hindi} & Label 1 & 0.43 & 0.74\\ 
& Label 2 & 0.59 & 0.73\\ 
& Label 3 & 0.70 & \textbf{0.81} \\ 
\hline
\multirow{3}{*}{Tamil} & Label 1 & 0.73 & 0.82\\ 
& Label 2 & 0.77 & 0.85\\ 
& Label 3 & 0.79 & \textbf{0.90}\\ 
\hline
\end{tabular}
\caption{F1 macro scores per label for fine-tuned models on our datasets in each language. Highest scores in each language are boldened.}
\label{tab:results}
\end{table}
\vspace{-2mm}
\section{Dataset Release}
Since assessment of gendered abuse is a subjective task, we are sharing the data with annotator level labels, instead of aggregate score based on the majority opinion ~\cite{prabhakaran-releasing}. We have anonymized the annotator names though they are recognized as authors on the paper. The data is shared under a CC BY 4.0 license as CSV files on GitHub.\footnote{\url{https://github.com/tattle-made/uli_dataset}}

\section{Model performance}
To assess performance of existing approaches to detect oGBV, we tested models from the automated abuse detection literature on our dataset. Specifically, we created train and test sets from annotations for Label 2 and fine-tuned models on them. We considered all data annotated by a single annotator as training data and ones annotated by multiple people (Table ~\ref{tab:annotation_number}) as test data, using majority labelling for the final label. For the models, we used IndicBERT~\citep{kakwani2020indicnlpsuite}, which is a multilingual ALBERT model trained on Indian language data, and XLM-T~\citep{barbieri-espinosaanke-camachocollados:2022:LREC}, a RoBERTa model fine-tuned on Twitter data, on our dataset. Our choice of model was motivated by the strong performance of IndicBERT over other multilingual models like mBERT and XLM-R when evaluated on tasks in Indian languages~\citep{kakwani2020indicnlpsuite}. For XLM-T, we relied on its pre-training on Twitter data and strong performance in prior social media based datasets as our primary motivation for inclusion. For all the models, we trained for 5 epochs with a learning rate of 5e-06, a batch size of 8 and the Adam optimizer. To avoid overfitting, we implement an early stopping mechanism conditioned on the evaluation F1 macro with a patience of 5 steps. We report the result in Table~\ref{tab:results}.

We see that the IndicBERT model is able to perform on Tamil fairly well. All three labels in English and Hindi Label 1 are the hardest for the model to learn. XLM-T, on the other hand, scores much better across the spectrum, which we hypothesize is due to its familiarity with Twitter data. Tamil is still the language with the highest performance, while English remains the hardest. This corresponds to the lower levels of agreement among the annotators for the English labels outlined in Table~\ref{tab:agreement}, demonstrating the subjectivity of the task.

\section{Discussion and Conclusion}
This paper presents an attempt to develop a dataset that centers the experience of those at the receiving end of gendered abuse, with their active participation. Through this dataset we seek to put into practice values in feminist and participatory AI such as inclusion, intersectionality, and co-designing systems with those who are subject to its decisions. The process and the resultant dataset surface numerous questions that need to be clarified through future work. First, we note a marked difference in the agreement scores in labels across languages. This could be a result of the difference in the posts selected in each language, or a difference in the interpretation of the annotation guideline by the annotators of each language. It could also be a result of difference in the diversity of annotator backgrounds in each language group. Understanding the source of heterogeneity in agreement scores in each language group needs further investigation. Second, a participatory project like this brings together people with different motivations. While all the annotators were motivated to address the challenge of online and offline gender based violence, the time they cold devote to the annotations varied. Availability of devices and familiarity with online interfaces to carry out the annotation also varied. The interplay of experts' motivations with the quality of annotations is a complex topic but one that needs attention when building participatory datasets and AI. Connected to motivations is the question of compensation. At present there is little guidance on compensation for experts' time in a project like this. Fair remuneration and recognition of expert contributions is an area of active research. Third, we recognize that oGBV is increasingly expressed through memes, images and videos. In future we hope to extend a similar approach to multi-modal content. Fourth, the process of creating this dataset was labor intensive. The core team that developed the annotation guideline comprised of people speaking the three languages. Such representation, however, may not have been feasible if we were working with ten languages. How best to balance the core goal of participatory design with material constraints of time and money is a question with non-obvious answers. Finally, while we relied on the majority vote on a label to test the ML models, we will continue to explore other approaches that don't flatten the disagreement across annotators.

\section{Limitations}
This work has a few limitations. Firstly, the data annotations solely concentrated on text-based abuse. The focus group discussions highlighted that a large proportion of abuse was in the form of images and videos. We hope to work on a similar approach to multi-modal content in the future. Secondly, the creation of our dataset was labor-intensive, prompting questions about managing participatory design goals with time and resource constraints. Lastly, there is a need for exploring approaches beyond majority voting to address disagreements among annotators in the dataset. In future work, we hope to circumvent some of these limitations and provide a more well-rounded approach to mitigating oGBV.

\section{Acknowledgments}
We would like to thank the funders who supported the project: Digital Society Challenge award granted to the Center for Internet and Society and Tattle Civic Tech; Mozilla Foundation's Data Futures Lab Cohort; and in-kind support by GitHub. At different stages of the project we consulted with experts in the field of gender studies, machine learning, content moderation and computational social science. The number of experts are too many to list here but we are grateful for their time and advice. 
%
 
\bibliography{anthology,custom}
\bibliographystyle{acl_natbib}

\appendix

\section{Appendix}
\label{sec:appendix}

\subsection{Annotator Disagreement}
Table \ref{tab:label1_2_disagreement} shows the number of posts where an annotator annotated label 1 and label 2 differently. The first two letters in the annotator ID describe the language the annotator was working with: English (en), Hindi (hi) or Tamil (ta). Comparisons about subjective opinions of annotators should not be drawn from this table since each annotator annotated a different number of posts. This data is not normalized. 

\subsection{Model links}
For reproducibility, we provide the models along with the corresponding Huggingface codes and links in Table~\ref{tab:model_links}

\begin{table}[h]
\begin{center}
   \begin{tabular}{ |p{20mm}|p{45mm}| } 
   \hline
    Model & Model Code \\
   \hline
    IndicBERT & \href{https://huggingface.co/ai4bharat/indic-bert}{ai4bharat/indic-bert} \\
    \hline
    BERT code-mixed & \href{https://huggingface.co/rohanrajpal/bert-base-en-hi-codemix-cased}{rohanrajpal/bert-base-en-hi-codemix-cased} \\
    \hline
    XLM-T & \href{https://huggingface.co/cardiffnlp/twitter-roberta-base-sep2021}{cardiffnlp/twitter-roberta-base-sep2021} \\
    \hline
  \end{tabular}
\end{center}
  \caption{Huggingface model codes for models used in the experiments}
  \label{tab:model_links}
\end{table}

\begin{table}[]
    \centering
    \begin{tabular}{|p{17mm}|p{17mm}|p{17mm}|p{17mm}| } 
    \hline
    Annotator ID & Posts with both & Posts with label 1:yes & Posts with label 1: no\\
                & labels marked yes  & label 2:no & and label 2:yes \\
    \hline
    en\_a1	&	82	&	5	&	92	\\
    \hline
    en\_a2	&	279	&	24	&	1	\\
    \hline
    en\_a3	&	809	&	4	&	142	\\
    \hline
    en\_a4	&	172	&	27	&	99	\\
    \hline
    en\_a5	&	411	&	4	&	547	\\
    \hline
    en\_a6	&	427	&	10	&	503	\\
    \hline
    hi\_a1	&	430	&	2	&	1151	\\
    \hline
    hi\_a2	&	334	&	66	&	485	\\
    \hline
    hi\_a3	&	713	&	412	&	600	\\
    \hline
    hi\_a4	&	239	&	110	&	264	\\
    \hline
    hi\_a5	&	670	&	9	&	637	\\
    \hline
    ta\_a1	&	955	&	2	&	679	\\
    \hline
    ta\_a2	&	848	&	9	&	512	\\
    \hline
    ta\_a4	&	1198	&	25	&	59	\\
    \hline
    ta\_a5	&	324	&	92	&	83	\\
    \hline
    ta\_a6	&	1075	&	234	&	29	\\
    \hline
    ta\_a7	&	245	&	0	&	73	\\
    
    \hline
    \end{tabular}
    \caption{Difference in opinion on label 1 and label 2 for all annotators}
    \label{tab:label1_2_disagreement}
\end{table}
\end{document}